\documentclass{IEEEtran}
\usepackage[absolute,showboxes,overlay]{textpos}
\usepackage{hyperref}

\TPMargin{5pt}

\newcommand{\copyrightstatement}{
    \begin{textblock}{0.84}(0.08,0.93)    
         \noindent
         \footnotesize
         \copyright 2022 IEEE. Personal use of this material is permitted. Permission from IEEE must be obtained for all other uses, in any current or future media, including reprinting/republishing this material for advertising or promotional purposes, creating new collective works, for resale or redistribution to servers or lists, or reuse of any copyrighted component of this work in other works. {Cite from IEEE. }\href{<https://doi.org/10.1109/ICASSP43922.2022.9746364>}{DOI No. 10.1109/ICASSP43922.2022.9746364}
    \end{textblock}
}

\usepackage{spconf,amsmath,graphicx}
\usepackage{epsfig}
\usepackage{amssymb}

\usepackage{cite}
\usepackage{amsfonts}
\usepackage{graphicx}
\usepackage{textcomp}
\usepackage{subfigure}
\usepackage{fancyhdr}
\usepackage{algorithm}
\usepackage{algpseudocode}
\usepackage{booktabs} 
\usepackage{multirow}
\usepackage{xcolor}

\title{Variational Bayesian Framework for Advanced Image Generation  \\
with Domain-Related Variables}
\name{
Yuxiao~Li\dag,
Santiago~Mazuelas\ddag, and
Yuan~Shen\dag}

\address{\dag
Department of Electronic Engineering,
Tsinghua University,
Beijing, China \\
\ddag BCAM-Basque Center for Applied Mathematics, Bilbao, Spain \\
Emails: li-yx18@mails.tsinghua.edu.cn,
smazuelas@bcamath.org,
shenyuan\_ee@tsinghua.edu.cn
}
\begin{document}

%
\maketitle
\copyrightstatement

\begin{abstract}

Deep generative models (DGMs) and their conditional counterparts provide a powerful ability for general-purpose generative modeling of data distributions. However, it remains challenging for existing methods to address advanced conditional generative problems without annotations, which can enable multiple applications like image-to-image translation and image editing. We present a unified Bayesian framework for such problems, which introduces an inference stage on latent variables within the learning process. In particular, we propose a variational Bayesian image translation network (VBITN) that enables multiple image translation and editing tasks. Comprehensive experiments show the effectiveness of our method on unsupervised image-to-image translation, and demonstrate the novel advanced capabilities for semantic editing and mixed domain translation.

\end{abstract}
\section{Introduction}
\label{sec:intro}
Deep generative models (DGMs) \cite{Goodfellow2014GenerativeAN,Kingma2014AutoEncodingVB} are popular ways to learn complicated data distributions in an unsupervised manner. 
However, they have less promising capabilities towards the generation of conditional distributions, and are hard to scale to different problems in a consistent scheme.
The related techniques include image-to-image translation and image editing, enabling a wide range of applications such as super resolution \cite{Ledig2017PhotoRealisticSI}, image colorization \cite{Zhang2017RealtimeUI}, image inpainting \cite{Yang2017HighResolutionII,Liu2018ImageIF}, and semantic attribute synthesis \cite{Park2019SemanticIS}. 

Different strategies have been proposed to improve the scalability of DGMs towards conditional distributions. Early conditional generative methods, like Conditional GAN \cite{Mirza2014ConditionalGA} and Info GAN \cite{Chen2016InfoGANIR}, use supervised annotations from the target distribution. While successful in basic conditional generative problems, these methods are insufficient towards advanced problems without direct annotations, such as unsupervised image-to-image generation. 
Existing techniques tackle this problem by adding constraints in either the image space or a low-dimensional latent space \cite{Zhu2017UnpairedIT,Liu2017UnsupervisedIT,liu2020gmm}.
However, a unified framework for the underlying generative process of different semantic variables is seldom claimed, resulting in redundant fine-tunning work and limited scalability towards advanced tasks in a consistent scheme.

From a statistical viewpoint, these problems can be described well by a latent variable model (LVM). Specifically, semantic features can be viewed as latent variables while the generation can be conducted by inferring the conditional distribution of images given the variables corresponding to desired semantics. The idea of disentangling codes for different semantics is partially discussed by \cite{Huang2018MultimodalUI,Park2020SwappingAF}, while seldom derived from first principles via statistic modeling.

In this paper, we present a novel probabilistic framework for a general class of conditional image generative problems. We then propose a deep generative network for image translation tasks, where latent variables of semantics are inferred via a variational lower bound in learning. Driven by a rigorous probabilistic model, the proposed method has a clear interpretation and improved generality to encompass multiple variants. Experimental results show that the proposed method achieves comparable performance with classic frameworks on unsupervised image-to-image translation, and enables novel variants like mixed domain translation.

\section{Bayesian Framework for Image Generation with latent variables}
\label{model}

We present a Bayesian framework for conditional image generation with respect to two latent variables, representing domain-related and domain-unrelated semantics respectively.

\subsection{Bayesian Model for Image Generation}
\label{sec:assumption}

Suppose the generative process of an image sample $\mathbf{x}^{(k)}\in\mathbb{X}$ in certain domain involves two latent variables: a domain-related variable $\mathbf{y}$ that describes features specific to the domain, and an independent domain-unrelated variable $\mathbf{z}$ that describes general features.
We refer to the domain-related variable as 'style' and the domain-unrelated variable as 'content', following the classical nomenclature in \cite{Gatys2016ImageST}.




\subsection{Unsupervised Image-to-Image Translation}




Consider a dataset $\mathbb{X}_S=\{\mathbf{x}_S^{(k)}\}_{k=1}^N$ consisting of $N$ i.i.d. samples of a random variable $\mathbf{x}_S$ corresponding with domain $S$, and a dataset $\mathbb{X}_T=\{\mathbf{x}_T^{(l)}\}_{l=1}^M$ consisting of $M$ i.i.d. samples of $\mathbf{x}_T$ corresponding with domain $T$. 
The content variables $\mathbf{z}_S$ and $\mathbf{z}_T$ corresponding with domains $S$ and $T$ share the same prior distribution $p(\mathbf{z})$, while the style variables $\mathbf{y}_S$ and $\mathbf{y}_T$ corresponding with these domains have different distributions, denoted as $p(\mathbf{y}_{S})$ and $p(\mathbf{y}_{T})$, respectively.

The translation process from an image $\mathbf{x}_S^{(k)}$ in domain $S$ to its counterpart $\mathbf{x}_{S\to T}^{(k)}$ in domain $T$, consists of three sequential steps: \textit{1)} A value $\textnormal{y}_{T}^{(k)}$ for style variable is generated from distribution $p(\mathbf{y}_{T})$ corresponding with domain $T$; \textit{2)} A value $\textnormal{z}_{S}^{(k)}$ for content variable is generated from the conditional distribution $p(\mathbf{z}|\mathbf{x}_S^{(k)})$; and \textit{3)} A translated image $\textnormal{x}^{(k)}_{S\to T}$ is generated from the conditional distribution $p(\mathbf{x}_T|\textnormal{y}_{T}^{(k)}, \textnormal{z}_{S}^{(k)})$.



\subsection{Multiple Variants}

The proposed model also enables to develop variants, achieved by modifications in the three steps above. We introduce three such variants, which can be further combined and varied.

\textbf{Multi-modal style editing.} 
The information for style semantics in the first step are obtained by sampling the distribution of the style variable, resulting in a spectrum of values. The translated image with these values can result in the generation of images with multi-modal styles, i.e.,
    \begin{equation}  \label{eq:style}
        \textnormal{y}_{T}^{(k_1)}, \ldots,\textnormal{y}_{T}^{(k_l)} \sim p(\mathbf{y}_{T}),
    \end{equation}
with the other steps stay unchanged, images of $l$ different styles in domain $T$ $\textnormal{x}_{S\to T}^{(k_1)}, \ldots, \textnormal{x}_{S\to T}^{(k_l)}$ can be generated.

\textbf{Multi-modal content editing.}
The information for content semantics in the second step are obtained by sampling the distribution of the content variable, resulting in a spectrum of values. The generation with these values can result in images of multi-modal contents, i.e.,
    \begin{equation}  \label{eq:content}
        \textnormal{z}_{S}^{(k_1)}, \ldots,\textnormal{z}_{S}^{(k_m)} \sim p(\mathbf{z}|\mathbf{x}_S^{(k)}),
    \end{equation}
with the other steps stay unchanged, images of $m$ content variants $\textnormal{x}_{S\to T}^{(k_1)}, \ldots, \textnormal{x}_{S\to T}^{(k_m)}$ can be achieved.

\textbf{Mixed domain translation.} 
The semantics determined by the style variable can represent a mixed style from more than one target domains, resulting in translated image in a mixed domain.
The distribution for the mixed style can be constructed as the weighted sum of style distributions, i.e.,
    \begin{equation}  \label{eq:mix}
        \textnormal{y}_{Mix}^{(k)} \sim p(\mathbf{y}_{Mix}) = \sum_{i=1}^n {w_i}p(\mathbf{y}_{T_n}), ~\sum_{i=1}^n w_i = 1,
    \end{equation}
where $w_i, i=1,\ldots,n$ are the weight values for these styles, e.g. $w_i=1/n, i=1,\ldots,n$. 



    
    
    
    
    

\section{Variational Bayesian Image Translation Network}
\label{sec:algorithm}

\begin{figure*}[ht]
    \vskip 0.2in
    \begin{center}
        \centerline{\includegraphics[width=0.8\textwidth]{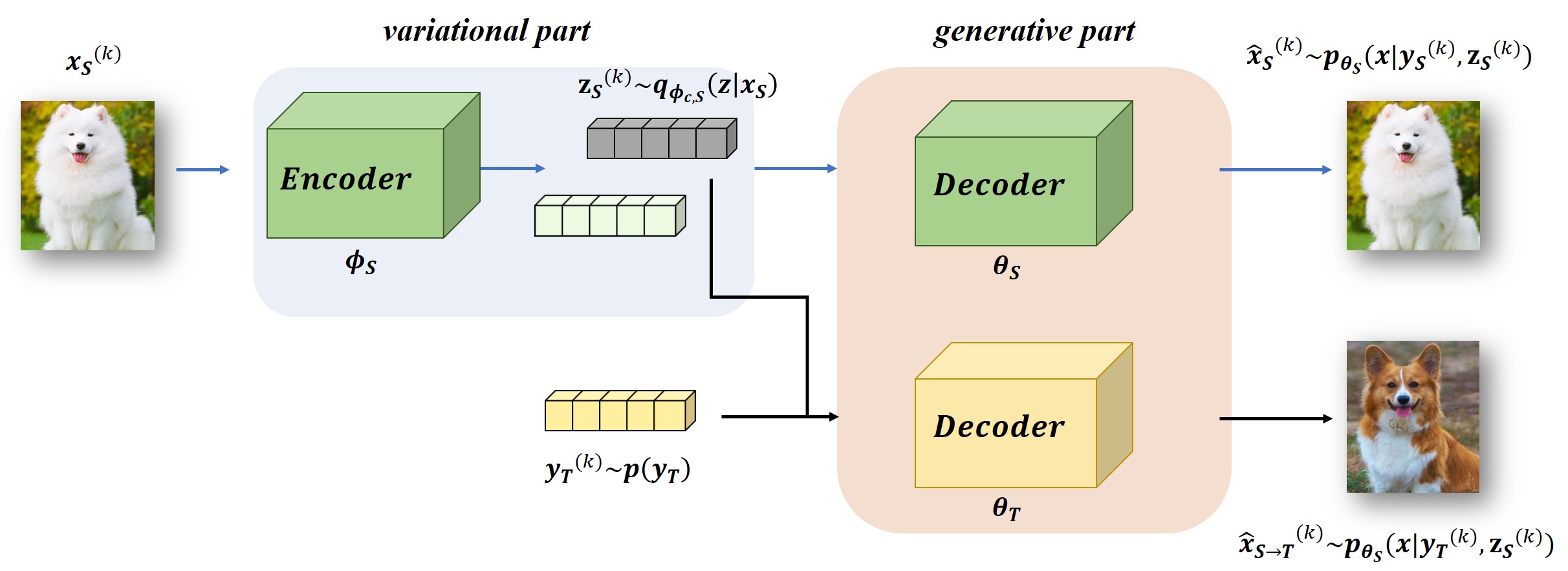}}
        \caption{The network architecture of the proposed VBITN. The framework consists of VAE-based networks which individually extract latent variables from different domain images. Then the learned latent variables are combined to generate new images.}
        \label{fig:network}
    \end{center}
    \vskip -0.2in
\end{figure*}

In this section, the variational Bayesian (VB) method is introduced and implemented to conduct image translation tasks. 

\subsection{Variational Bayesian Method}
\label{sec:vb}

According to the VB technique, we construct a distribution $q(\mathbf{y}, \mathbf{z}|\mathbf{x})$ to approximate the true posterior $p(\mathbf{y}, \mathbf{z}|\mathbf{x})$. 
Following the mean field approximation, we assume that
\begin{equation}  \label{eq:independence}
    q(\mathbf{y}, \mathbf{z}|\mathbf{x}) = q(\mathbf{y}|\mathbf{x})q(\mathbf{z}|\mathbf{x}).
\end{equation}

The following proposition shows the lower bound of the log-likelihood $\log p(\mathbf{x})$ of each sample in the LVM.

\textbf{Proposition 1.} 
Given the Bayesian model for image translation and the variational distribution $q$ on latent variables, the evidence lower bound (ELBO) $\mathcal{L}$ of log-likelihood $\log p(\mathbf{x})$ is expressed as follows, 
\begin{equation}  \label{eq:bound}
    \begin{aligned}
        \mathcal{L}(p, q;\mathbf{x}) =& \mathbb{E}_{q(\mathbf{y},\mathbf{z}|\mathbf{x})}\big[\log p(\mathbf{x}|\mathbf{y},\mathbf{z})\big]  \\
        &- \operatorname{KL}\big(q(\mathbf{y}|\mathbf{x}) \big|\big| p(\mathbf{y})\big) - \operatorname{KL}\big(q(\mathbf{z}|\mathbf{x}) \big|\big| p(\mathbf{z})\big)
    \end{aligned}
\end{equation}
\noindent where $\operatorname{KL}$ is the Kullback-Leibler (KL) divergence.

Such bound can be used to find a suitable approximated distribution $q^*$ that matches the true distribution $p$ in general. 

\subsection{Neural Modules for Distributions}

We construct neural modules to represent the unknown $p$ and the variational distribution $q$. In particular, the likelihood distribution is assumed to come from a parametric family $p_{\boldsymbol{\theta}}(\mathbf{x}|\mathbf{y},\mathbf{z})$ learned by a decoder network $g_{\boldsymbol{\theta}}$, while the variational posterior distribution is from $q_{\boldsymbol{\phi}}(\mathbf{y},\mathbf{z}|\mathbf{x})$ learned by an encoder network $f_{\boldsymbol{\phi}}$. 
Combing the neural modules, we can construct a modified VAE to learn variational parameter $\boldsymbol{\phi}$ jointly with the likelihood parameter $\boldsymbol{\theta}$ via the lower bound.

The likelihood distribution is from the parametric family, learned by the decoder network $g_{\boldsymbol{\theta}}$ as follows,
\begin{equation}
    \textnormal{x}^{(k)}=g_{\boldsymbol{\theta}}(\textnormal{y}^{(k)}, \textnormal{z}^{(k)}) \sim p_{\boldsymbol{\theta}}(\mathbf{x}|\mathbf{y}^{(k)}, \mathbf{z}^{(k)})
\end{equation}

According to the properties of the latent variables, we assume the prior distributions are as follows:
\begin{equation}  \label{eq:priors}
    p(\mathbf{y})=\mathcal{N}(\boldsymbol{\alpha}, \mathbf{I}),~p(\mathbf{z})=\mathcal{N}(\boldsymbol{0}, \mathbf{I})
\end{equation}
where $\mathbf{y}, \boldsymbol{\alpha}\in\mathbb{R}^{D_s}, \mathbf{z}\in\mathbb{R}^{D_c}$ and $\boldsymbol{\alpha}$ is domain-related. 
The choices of domain parameter $\boldsymbol{\alpha}_S, \boldsymbol{\alpha}_T$ can be arbitrary as long as $\boldsymbol{\alpha}_S\neq\boldsymbol{\alpha}_T$.

The approximated posterior distributions in this case are also Gaussian with learned parameters by network $f_{\boldsymbol{\phi}}$,
\begin{equation}  \label{eq:decomposeq}
      \begin{aligned}
         q_{\boldsymbol{\phi}_{s}}(\mathbf{y}|\mathbf{x})=\mathcal{N}(\mathbf{y};\hat{\boldsymbol{\mu}}_{s}, \hat{\boldsymbol{\sigma}}^2_{s}\boldsymbol{I}),  q_{\boldsymbol{\phi}_{c}}(\mathbf{z}|\mathbf{x})=\mathcal{N}(\mathbf{z};\hat{\boldsymbol{\mu}}_{c}, \hat{\boldsymbol{\sigma}}^2_{c}\boldsymbol{I})
      \end{aligned}
\end{equation}
where $\hat{\boldsymbol{\mu}}_{s}, \hat{\boldsymbol{\sigma}}^2_{s} \in \mathbb{R}^{D_s}$ and $\hat{\boldsymbol{\mu}}_{c}, \hat{\boldsymbol{\sigma}}^2_{c} \in \mathbb{R}^{D_c}$.

\subsection{Parametric Form of ELBO}
\label{sec:bound}

To utilize the gradient descent algorithm for network learning, we derive the analytical version of the variational lower bound with respect to parameters $\boldsymbol{\phi}$ and $\boldsymbol{\theta}$, expressed as follows,
\begin{equation}  \label{eq:bound_para}
    \begin{aligned}
        \mathcal{L}(\boldsymbol{\phi}, \boldsymbol{\theta};\mathbf{x}) &= \mathbb{E}_{q_{\boldsymbol{\phi}}(\mathbf{y},\mathbf{z}|\mathbf{x})}\big[\log p_{\boldsymbol{\theta}}(\mathbf{x}|\mathbf{y},\mathbf{z})\big]  \\
        &- \operatorname{KL}\big(q_{\boldsymbol{\phi}}(\mathbf{y}|\mathbf{x}) \big|\big| p(\mathbf{y})\big) - \operatorname{KL}\big(q_{\boldsymbol{\phi}}(\mathbf{z}|\mathbf{x}) \big|\big| p(\mathbf{z})\big)
    \end{aligned}
\end{equation}
  
The last two terms can be integrated analytically with the Gaussian assumptions. The first term is evaluated as follows,
  \begin{equation}
      \mathbb{E}_{q_{\boldsymbol{\phi}}(\mathbf{y},\mathbf{z}|\mathbf{x})}\big[\log p_{\boldsymbol{\theta}}(\mathbf{x}|\mathbf{y},\mathbf{z})\big]
          = \frac{1}{L}\sum_{l=1}^L \log p_{\boldsymbol{\theta}}(\mathbf{x}|\mathbf{y}^{(l)},\mathbf{z}^{(l)})
  \end{equation}
where $\boldsymbol{\epsilon}^{(l)}\sim\mathcal{N}(\boldsymbol{0},\boldsymbol{I})$, $\mathbf{y}^{(l)}= \hat{\boldsymbol{\mu}}_{s} + \hat{\boldsymbol{{\sigma}}}^2_{s} \odot \boldsymbol{\epsilon}^{(l)}$, $\mathbf{z}^{(l)}= \hat{\boldsymbol{\mu}}_{c} + \hat{\boldsymbol{{\sigma}}}^2_{c} \odot \boldsymbol{\epsilon}^{(l)}$ using the so-called reparameterization trick \cite{Kingma2014AutoEncodingVB}.

\subsection{Network Learning}


Suppose we are given a dataset $\mathbb{X}_S$ from the source domain, and $N$ unpaired datasets $\{\mathbb{X}_{T_i}\}_{i=1}^N$ from the target domains. The target is to translate some sample $\mathbf{x}^{(k)}_S$ from domain $S$ to its counterpart with the mixed style of the regarding target domains. 
We adopt a compound loss with three terms: an inter-domain loss $\mathbb{L}_{\text{ind}}$ for latent variables inference, an adversarial loss $\mathbb{L}_{\text{adv}}$ to enforce realism of the translated images,
and a reconstruction loss $\mathbb{L}_{\text{rec}}$ in for latent variable regularization.

Denote $\boldsymbol{\phi}_S$ and $\boldsymbol{\theta}_S$ for parameters of domain $S$, while $\boldsymbol{\phi}_{T_i}$ and $\boldsymbol{\theta}_{T_i}$ for domain $T_i$.
We first implement the inter-domain loss $\mathbb{L}_{\text{ind}}$ as the expectation of negative inter-domain bounds on corresponding datasets:
\begin{equation}
    \begin{aligned}
        \mathbb{L}_{\text{ind}} =& \mathbb{E}_{\mathbb{X}_S}\big[{\mathcal{L}}(\boldsymbol{\theta}_S, \boldsymbol{\phi}_S; \mathbf{x})\big] + \sum_{i=1}^N \mathbb{E}_{\mathbb{X}_{T_i}}\big[{\mathcal{L}}(\boldsymbol{\theta}_{T_i}, \boldsymbol{\phi}_{T_i}; \mathbf{x})\big].
    \end{aligned}
\end{equation}

The next two terms, $\mathbb{L}_{\text{rec}}$ and $\mathbb{L}_{\text{adv}}$ are constructed to form regularization in both latent space and image space to constraint learning, expressed as follows, 
    \begin{equation}
        \begin{aligned}
         \mathbb{L}_{\text{rec}} =& \sum_{i=1}^N \mathbb{E}_{\mathbb{X}_{S\to T_i}}\mathbb{E}_{q_{\boldsymbol{\phi}_{S}}(\mathbf{z}|\mathbf{x})}\big[ \Vert \mathbf{z} - \mathbf{z}_{S} \Vert^2 \big]  \\
        &\quad\quad+ \mathbb{E}_{\sim \mathbb{X}_{S\to T_i}} \mathbb{E}_{q_{\boldsymbol{\phi}_{T_i}}(\mathbf{y}|\mathbf{x})}\big[\Vert \mathbf{y} - \mathbf{y}_{T_i} \Vert^2\big]
        \end{aligned}
    \end{equation}

\begin{equation}
    \begin{aligned}
        \mathbb{L}_{\text{adv}} =&\sum_{i=1}^N \mathbb{E}_{{\mathbb{X}}_{T_i}}\Big[\log \big(1-D_{\boldsymbol{\varphi}}\big(\mathbf{x}\big)\big)\Big]  \\
        &\quad\quad +  \mathbb{E}_{{\mathbb{X}}_{S\to T_i}}\big[\log D_{\boldsymbol{\varphi}}(\mathbf{x})\big]
    \end{aligned}
\end{equation}
where $D_{\boldsymbol{\varphi}}(\cdot)$ denotes the discriminator network with parameter $\boldsymbol{\varphi}$ to distinguish between true and generated images.

\begin{table*}[t]
\caption{LPIPS \cite{Zhang2018TheUE} and AMT \cite{Zhang2016ColorfulIC} scores for different methods on unsupervised image-to-image translation on dataset 'Monet's painting$\leftrightarrow$Photo'. The best two results are
highlighted in red and blue colors respectively.}
\label{tab:exp_comp}
\begin{center}
\begin{small}
\begin{sc}
\begin{tabular}{lccccr}
\toprule
\multirow{2}{*}{Method} & \multicolumn{2}{c}{Photo$\rightarrow$Monet's Painting} & \multicolumn{2}{c}{Monet's Painting$\rightarrow$Photo} \\
& LPIPS (Diversity) & AMT (Realism) & LPIPS (Diversity) & AMT (Realism) \\
\midrule
CycleGAN \cite{Zhu2017UnpairedIT}    &.6705$\pm$ .0025& \textcolor{blue}{37.28$\pm$2.26\%} &.6604$\pm$ .0031& 17.58$\pm$2.24\% \\
BicycleGAN \cite{Zhu2017TowardMI} & .5982$\pm$ .0026& 19.31$\pm$1.89\% & .5805$\pm$ .0026& 15.46$\pm$2.43\% \\
DiscoGAN \cite{Kim2017LearningTD}    & .6775$\pm$.0026 & 31.49$\pm$2.67\% & .6667$\pm$ .0027& \textcolor{blue}{24.43$\pm$3.01\%} \\
DualGAN \cite{Yi2017DualGANUD}    & \textcolor{blue}{.6957$\pm$.0029} & 15.84$\pm$2.28\% & \textcolor{red}{.7012$\pm$.0030} & 19.29$\pm$2.13\% \\
UNIT \cite{Liu2017UnsupervisedIT}     & .6734$\pm$ .0026& 34.22$\pm$2.46\% & .6661$\pm$ .0024& 21.43$\pm$1.89\% \\
MUNIT \cite{Huang2018MultimodalUI}      & .4544$\pm$ .0028& 17.86$\pm$2.89\% & .6536$\pm$ .0027& 13.85$\pm$2.75\%  \\
VBITN (Ours)      & \textcolor{red}{.6997$\pm$ .0024} & \textcolor{red}{38.62$\pm$2.24\%} & \textcolor{blue}{.6725$\pm$ .0022}& \textcolor{red}{27.30$\pm$1.87\%}  \\
\bottomrule
\end{tabular}
\end{sc}
\end{small}
\end{center}
\vskip -0.1in
\end{table*}

\begin{figure*}[!ht]
    \vskip 0.2in
    \begin{center}
        \subfigure[]{
        \begin{minipage}[t]{0.95\linewidth}
        \centerline{\includegraphics[width=0.95\textwidth]{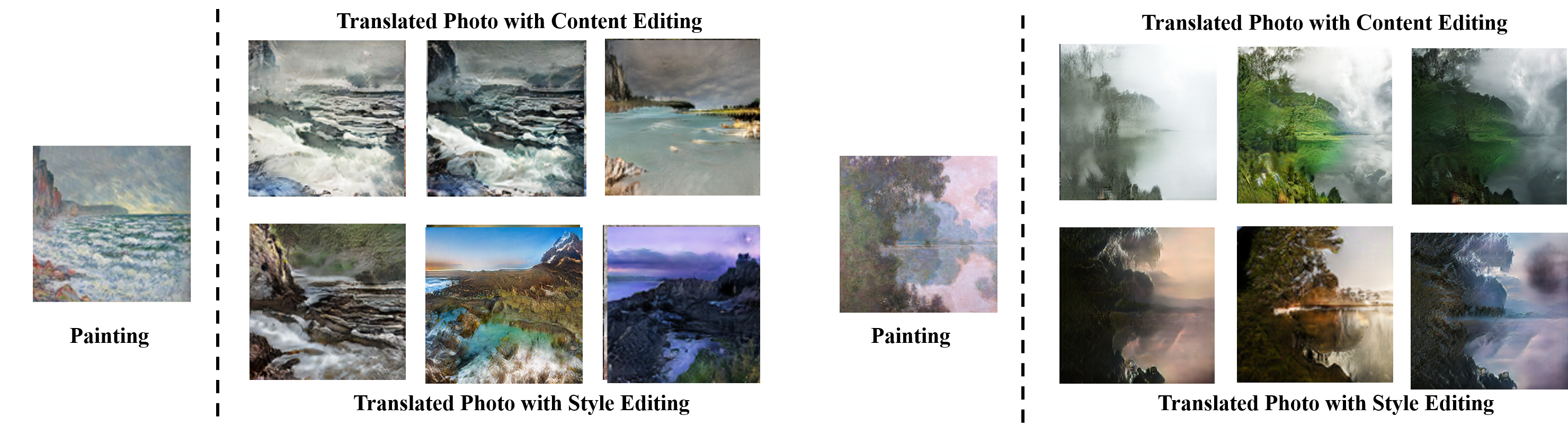}}
        \end{minipage}%
        }  \\
        \vskip -0.05in
        \subfigure[]{
        \begin{minipage}[t]{0.95\linewidth}
        \centerline{\includegraphics[width=0.95\textwidth]{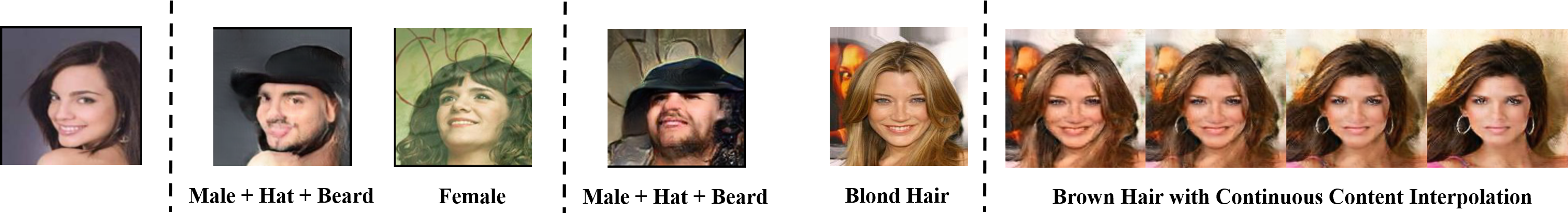}}
        \end{minipage}
        }
        \vskip -0.05in
        \caption{VBITN enables efficient unsupervised image-to-image translation as well as semantic editing and mixed domain translation: (a) Paintings are translated to photos with different semantics; (b) Mixed domain translation on human face attributes.}
        \label{fig:exp_edit}
    \end{center}
    \vskip -0.2in
\end{figure*}

\section{Experiments}
\label{sec:exp}

\subsection{Experimental Setup}

We compare our Bayesian framework on image translation task with several classic methods, including Cycle GAN \cite{Zhu2017UnpairedIT}, Bicycle GAN \cite{Zhu2017TowardMI}, Disco GAN \cite{Kim2017LearningTD}, Dual GAN \cite{Yi2017DualGANUD} utilizing cycle-consistency techniques, and UNIT \cite{Liu2017UnsupervisedIT} and MUNIT \cite{Huang2018MultimodalUI} utilizing latent representation techniques\footnote{Note that only classic framework-level methods for the basic image-to-image translation are adopted as baselines. More advanced works like StyleGAN\cite{Karras2020AnalyzingAI,Karras2019ASG} and StarGAN\cite{Choi2018StarGANUG} are not compared and can be viewed as implementable techniques on any basic frameworks.}.

We evaluate our techniques on the 'Monet's painting  $\leftrightarrow$ photo' dataset and CelebA dataset \cite{Liu2015DeepLF}, all at resolution of $128$\emph{px}. Quantitative comparisons with related methods are conducted by the Learned Perceptual Image Patch Similarity (LPIPS) distance \cite{Zhang2018TheUE} for diversity, and Amazon Mechanical Turk (AMT) perceptual \cite{Zhang2016ColorfulIC} for realism, claimed sufficient in other literature. 

\subsection{Unsupervised Image-to-Image Translation}
\label{sec:common_exp}


Table~\ref{tab:exp_comp} reports the achieved performance different methods on LPIPS metric and AMT studies. We observe that our competitors tend to suffer from a trade-off between diversity and realism, though achieve remarkable results in one of the metrics.
Our method gets the best of both sides, as it encourages diverse outputs with semantic variables and also has a well-defined objective function for regularization.

\subsection{Multiple Variants}
\label{sec:variant_exp}

Qualitative results of our method on semantic editing are shown in Figure~\ref{fig:exp_edit}(a). 
We observe that both content and style semantics of the generated image can have meaningful variants with little cost to quality. Figure~\ref{fig:exp_edit}(b) shows our test on the novel mixed domain translation. The domain-related attributes (style) 'male', 'hat' and 'beard' have been successfully translated, while the domain-unrelated attributes (content) like 'looks' and 'expressions' are randomly sampled. Our method can produce translated image with multiple attributes with sharp edges and reliable details.

\section{Conclusion}

We introduced a Bayesian framework for conditional generative problems, and proposed VBITN for related tasks. 
The contributions include regularizing the ill-posed nature of image translation, and enabling novel capabilities like semantic editing. 
The developed techniques also suggest potential of combining DGMs and statistic tools to develop inference ability. 
Future work will tackle more scalable frameworks via delicate designs in latent space and graphic model.

\vfill\pagebreak

\bibliographystyle{IEEEtran}
\bibliography{refs}

\begin{thebibliography}{10}
\providecommand{\url}[1]{#1}
\csname url@samestyle\endcsname
\providecommand{\newblock}{\relax}
\providecommand{\bibinfo}[2]{#2}
\providecommand{\BIBentrySTDinterwordspacing}{\spaceskip=0pt\relax}
\providecommand{\BIBentryALTinterwordstretchfactor}{4}
\providecommand{\BIBentryALTinterwordspacing}{\spaceskip=\fontdimen2\font plus
\BIBentryALTinterwordstretchfactor\fontdimen3\font minus
  \fontdimen4\font\relax}
\providecommand{\BIBforeignlanguage}[2]{{%
\expandafter\ifx\csname l@#1\endcsname\relax
\typeout{** WARNING: IEEEtran.bst: No hyphenation pattern has been}%
\typeout{** loaded for the language `#1'. Using the pattern for}%
\typeout{** the default language instead.}%
\else
\language=\csname l@#1\endcsname
\fi
#2}}
\providecommand{\BIBdecl}{\relax}
\BIBdecl

\bibitem{Goodfellow2014GenerativeAN}
I.~J. Goodfellow, J.~Pouget-Abadie, M.~Mirza, B.~Xu, D.~Warde-Farley, S.~Ozair,
  A.~C. Courville, and Y.~Bengio, ``Generative adversarial nets,'' in
  \emph{NIPS}, 2014.

\bibitem{Kingma2014AutoEncodingVB}
D.~P. Kingma and M.~Welling, ``Auto-encoding variational bayes,'' \emph{CoRR},
  vol. abs/1312.6114, 2014.

\bibitem{Ledig2017PhotoRealisticSI}
C.~Ledig, L.~Theis, F.~Husz{\'a}r, J.~Caballero, A.~Aitken, A.~Tejani, J.~Totz,
  Z.~Wang, and W.~Shi, ``Photo-realistic single image super-resolution using a
  generative adversarial network,'' \emph{2017 IEEE Conference on Computer
  Vision and Pattern Recognition (CVPR)}, pp. 105--114, 2017.

\bibitem{Zhang2017RealtimeUI}
R.~Zhang, J.-Y. Zhu, P.~Isola, X.~Geng, A.~Lin, T.~Yu, and A.~A. Efros,
  ``Real-time user-guided image colorization with learned deep priors,''
  \emph{ACM Trans. Graph.}, vol.~36, pp. 119:1--119:11, 2017.

\bibitem{Yang2017HighResolutionII}
C.~Yang, X.~Lu, Z.~L. Lin, E.~Shechtman, O.~Wang, and H.~Li, ``High-resolution
  image inpainting using multi-scale neural patch synthesis,'' \emph{2017 IEEE
  Conference on Computer Vision and Pattern Recognition (CVPR)}, pp.
  4076--4084, 2017.

\bibitem{Liu2018ImageIF}
G.~Liu, F.~Reda, K.~Shih, T.~Wang, A.~Tao, and B.~Catanzaro, ``Image inpainting
  for irregular holes using partial convolutions,'' \emph{ArXiv}, vol.
  abs/1804.07723, 2018.

\bibitem{Park2019SemanticIS}
T.~Park, M.-Y. Liu, T.~Wang, and J.-Y. Zhu, ``Semantic image synthesis with
  spatially-adaptive normalization,'' \emph{2019 IEEE/CVF Conference on
  Computer Vision and Pattern Recognition (CVPR)}, pp. 2332--2341, 2019.

\bibitem{Mirza2014ConditionalGA}
M.~Mirza and S.~Osindero, ``Conditional generative adversarial nets,''
  \emph{ArXiv}, vol. abs/1411.1784, 2014.

\bibitem{Chen2016InfoGANIR}
X.~Chen, Y.~Duan, R.~Houthooft, J.~Schulman, I.~Sutskever, and P.~Abbeel,
  ``Infogan: Interpretable representation learning by information maximizing
  generative adversarial nets,'' in \emph{NIPS}, 2016.

\bibitem{Zhu2017UnpairedIT}
J.-Y. Zhu, T.~Park, P.~Isola, and A.~A. Efros, ``Unpaired image-to-image
  translation using cycle-consistent adversarial networks,'' \emph{2017 IEEE
  International Conference on Computer Vision (ICCV)}, pp. 2242--2251, 2017.

\bibitem{Liu2017UnsupervisedIT}
M.-Y. Liu, T.~Breuel, and J.~Kautz, ``Unsupervised image-to-image translation
  networks,'' in \emph{NIPS}, 2017.

\bibitem{liu2020gmm}
Y.~Liu, M.~De~Nadai, J.~Yao, N.~Sebe, B.~Lepri, and X.~Alameda-Pineda,
  ``Gmm-unit: Unsupervised multi-domain and multi-modal image-to-image
  translation via attribute gaussian mixture modeling,'' \emph{arXiv preprint
  arXiv:2003.06788}, 2020.

\bibitem{Huang2018MultimodalUI}
X.~Huang, M.-Y. Liu, S.~J. Belongie, and J.~Kautz, ``Multimodal unsupervised
  image-to-image translation,'' in \emph{ECCV}, 2018.

\bibitem{Park2020SwappingAF}
T.~Park, J.-Y. Zhu, O.~Wang, J.~Lu, E.~Shechtman, A.~A. Efros, and R.~Zhang,
  ``Swapping autoencoder for deep image manipulation,'' \emph{ArXiv}, vol.
  abs/2007.00653, 2020.

\bibitem{Gatys2016ImageST}
L.~A. Gatys, A.~S. Ecker, and M.~Bethge, ``Image style transfer using
  convolutional neural networks,'' \emph{2016 IEEE Conference on Computer
  Vision and Pattern Recognition (CVPR)}, pp. 2414--2423, 2016.

\bibitem{Zhang2018TheUE}
R.~Zhang, P.~Isola, A.~A. Efros, E.~Shechtman, and O.~Wang, ``The unreasonable
  effectiveness of deep features as a perceptual metric,'' \emph{2018 IEEE/CVF
  Conference on Computer Vision and Pattern Recognition}, pp. 586--595, 2018.

\bibitem{Zhang2016ColorfulIC}
R.~Zhang, P.~Isola, and A.~A. Efros, ``Colorful image colorization,'' in
  \emph{ECCV}, 2016.

\bibitem{Zhu2017TowardMI}
J.-Y. Zhu, R.~Zhang, D.~Pathak, T.~Darrell, A.~A. Efros, O.~Wang, and
  E.~Shechtman, ``Toward multimodal image-to-image translation,'' in
  \emph{NIPS}, 2017.

\bibitem{Kim2017LearningTD}
T.~Kim, M.~Cha, H.~Kim, J.~K. Lee, and J.~Kim, ``Learning to discover
  cross-domain relations with generative adversarial networks,'' \emph{ArXiv},
  vol. abs/1703.05192, 2017.

\bibitem{Yi2017DualGANUD}
Z.~Yi, H.~Zhang, P.~Tan, and M.~Gong, ``Dualgan: Unsupervised dual learning for
  image-to-image translation,'' \emph{2017 IEEE International Conference on
  Computer Vision (ICCV)}, pp. 2868--2876, 2017.

\bibitem{Karras2020AnalyzingAI}
T.~Karras, S.~Laine, M.~Aittala, J.~Hellsten, J.~Lehtinen, and T.~Aila,
  ``Analyzing and improving the image quality of stylegan,'' \emph{2020
  IEEE/CVF Conference on Computer Vision and Pattern Recognition (CVPR)}, pp.
  8107--8116, 2020.

\bibitem{Karras2019ASG}
T.~Karras, S.~Laine, and T.~Aila, ``A style-based generator architecture for
  generative adversarial networks,'' \emph{2019 IEEE/CVF Conference on Computer
  Vision and Pattern Recognition (CVPR)}, pp. 4396--4405, 2019.

\bibitem{Choi2018StarGANUG}
Y.~Choi, M.-J. Choi, M.~Kim, J.-W. Ha, S.~Kim, and J.~Choo, ``Stargan: Unified
  generative adversarial networks for multi-domain image-to-image
  translation,'' \emph{2018 IEEE/CVF Conference on Computer Vision and Pattern
  Recognition}, pp. 8789--8797, 2018.

\bibitem{Liu2015DeepLF}
Z.~Liu, P.~Luo, X.~Wang, and X.~Tang, ``Deep learning face attributes in the
  wild,'' \emph{2015 IEEE International Conference on Computer Vision (ICCV)},
  pp. 3730--3738, 2015.

\end{thebibliography}

\end{document}